%% file: main.tex
\documentclass{article} 
\usepackage{iclr2026_re-align_workshop,times}

\input{math_commands.tex}

\usepackage{hyperref}
\usepackage{url}
\usepackage{graphicx}
\usepackage{booktabs}
\usepackage{multirow}
\usepackage{float}

\title{ASPEN: Spectral-Temporal Fusion \\
for Cross-Subject Brain Decoding\thanks{Code available at \url{https://github.com/meganmlee/ASPEN}}}

\author{
  Megan Lee\textsuperscript{1} \quad
  Seung Ha Hwang\textsuperscript{1,2} \quad
  Inhyeok Choi\textsuperscript{1,3} \quad
  Shreyas Darade\textsuperscript{1} \And
  Mengchun Zhang\textsuperscript{4} \quad
  Kateryna Shapovalenko\textsuperscript{1} \\[1em]
  \textsuperscript{1}Carnegie Mellon University, Pittsburgh, PA, USA \\
  \textsuperscript{2}Kyung Hee University, Yongin, South Korea \\
  \textsuperscript{3}Korea Advanced Institute of Science and Technology, Deajeon, South Korea \\
  \textsuperscript{4}University of Pittsburgh, Pittsburgh, PA, USA \\
}

\iclrfinalcopy 
\begin{document}

\maketitle

\fancypagestyle{firstpage}{
  \fancyfoot[L]{\textit{Preprint.}}
  \fancyfoot[C]{}
  \fancyfoot[R]{}
}
\thispagestyle{firstpage}

\begin{abstract}
Cross-subject generalization in EEG-based brain-computer interfaces (BCIs) remains challenging due to individual variability in neural signals. We investigate whether spectral representations offer more stable features for cross-subject transfer than temporal waveforms. Through correlation analyses across three EEG paradigms (SSVEP, P300, and Motor Imagery), we find that spectral features exhibit consistently higher cross-subject similarity than temporal signals. Motivated by this observation, we introduce ASPEN, a hybrid architecture that combines spectral and temporal feature streams via multiplicative fusion, requiring cross-modal agreement for features to propagate. Experiments across six benchmark datasets reveal that ASPEN is able to dynamically achieve the optimal spectral-temporal balance depending on the paradigm. ASPEN achieves the best unseen-subject accuracy on three of six datasets and competitive performance on others, demonstrating that multiplicative multimodal fusion enables effective cross-subject generalization.
\end{abstract}

\section{Introduction}

Cross-subject generalization remains a fundamental bottleneck in EEG-based brain-computer interfaces (BCIs). Models trained on multi-subject data often degrade substantially when deployed to new users, requiring lengthy subject-specific calibration that undermines the goal of plug-and-play systems \citep{wan2021review, liang2024adaptive}. This is due to inherent differences between individuals such as skull thickness, cortical folding, and electrode placement that can produce substantial variation in signal amplitude, timing, and spatial distribution \citep{lu2024domain, roy2019deep}.

A growing body of work has addressed this limitation through increasingly expressive temporal modeling, progressing from compact CNN-based decoders \citep{lawhern2018eegnet} to Transformer architectures that capture global dependencies \citep{song2022eeg}. However, temporal waveforms are highly sensitive to phase shifts, latency jitter, and amplitude scaling across subjects. The hypothesis we investigate here is that spectral representations provide a more stable basis for cross-subject transfer. Frequency-domain features abstract away precise timing information while preserving the oscillatory signatures, such as $\mu$ (8--12 Hz) and $\beta$ (13--30 Hz) rhythms, that serve as primary biomarkers for BCI paradigms \citep{ang2008fbcsp, mane2020multi}.

To test this hypothesis, we first conduct a systematic correlation analyses comparing temporal and spectral representations across SSVEP, P300, and Motor Imagery paradigms. Our analysis reveals that spectral features exhibit substantially higher cross-subject similarity than temporal signals, suggesting that frequency-domain representations offer a more robust foundation for generalization.
\begin{figure} [h]	
    \centering
    \includegraphics[width=1\linewidth]{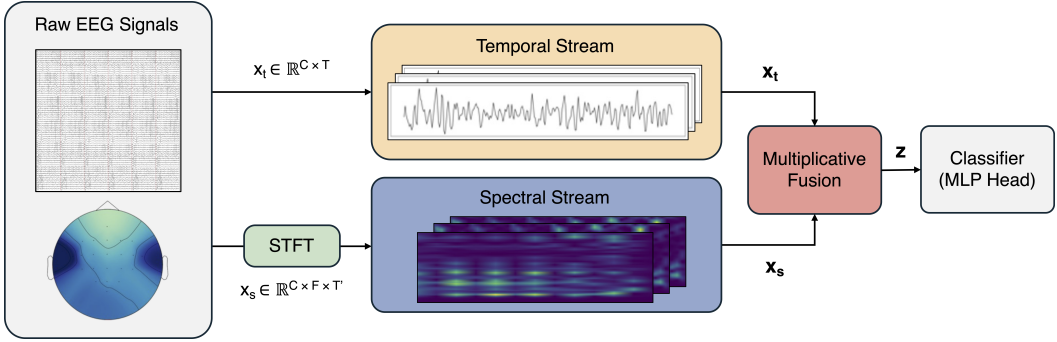}
    \caption{ASPEN architecture. Raw EEG is processed through parallel temporal and spectral streams, combined via multiplicative fusion before classification.}
    \label{fig:model}
    \vspace{-10pt}
\end{figure}
Motivated by this finding, we introduce ASPEN (\textbf{A}daptive \textbf{Sp}ectral \textbf{E}ncoder \textbf{N}etwork, Figure \ref{fig:model}), a hybrid framework that processes EEG signals through parallel temporal and spectral streams and combines them via multiplicative fusion. Unlike prior approaches that concatenate or average multimodal features \citep{li2021tssenet, li2025tsformer}, multiplicative fusion computes element-wise products of projected stream representations, requiring both streams to agree for a feature to propagate. This cross-modal gating naturally suppresses artifacts and noise that appear prominently in only one view, while amplifying genuine neural patterns that manifest consistently across both temporal and spectral domains.

We evaluate ASPEN across six benchmark datasets spanning three paradigms. Our experiments reveal that the optimal spectral-temporal balance varies by task: P300 decoding benefits strongly from spectral emphasis, while Motor Imagery requires greater temporal contribution. ASPEN achieves the best unseen-subject accuracy on three datasets (Lee2019 SSVEP, BNCI2014 P300, and Lee2019 MI), outperforming both specialized temporal models and recent multimodal transformers. These results demonstrate that our model is able to perform cross-subject generalization across different BCI tasks while maintaining robustness across diverse neural signatures.

\subsection{Related Work}

\textbf{Temporal modeling:} Deep learning for EEG signals has evolved from high-capacity architectures like DeepConvNet \citep{schirrmeister2017deep} toward compact, neurophysiologically-informed models. EEGNet \citep{lawhern2018eegnet} introduced depthwise and separable convolutions that mirror traditional spatial filtering, achieving strong performance with minimal parameters. Transformer-based models such as EEG Conformer \citep{song2022eeg} and hybrid CNN-Transformer architectures like CTNet \citep{zhao2024ctnet} capture long-range temporal dependencies. Temporal convolutional networks (TCNs) offer improved sequential modeling with training stability advantages over recurrent approaches \citep{ingolfsson2020eeg, musallam2021tcn}.

\textbf{Spectral and filter-bank approaches:} Filter-bank methods decompose EEG into frequency sub-bands before learning spatial filters. The foundational FBCSP algorithm \citep{ang2008fbcsp} demonstrated that isolating discriminative frequency bands improves motor imagery classification. Deep learning extensions apply this principle with learnable filters \citep{mane2020multi, liu2022fbmsnet}, while IFNet \citep{wang2023ifnet} models cross-frequency interactions. Time-frequency representations via wavelets have also shown promise for capturing non-stationary dynamics \citep{ morales2022timefreq}.

\textbf{Multimodal fusion:} Recent work has begun combining temporal and spectral features. Li et al. \citep{li2021tssenet} proposed a temporal-spectral squeeze-and-excitation network for motor imagery. TSformer-SA \citep{li2025tsformer} integrates temporal signals with wavelet spectrograms through cross-view attention for RSVP decoding. Dual-branch architectures have also been explored for emotion recognition \citep{luo2023dual}. However, these approaches typically employ additive fusion strategies like concatenation, averaging, or learned weighted sums that allow each stream to contribute independently. Our multiplicative approach instead enforces cross-modal agreement, acting as a feature-wise AND gate that filters unreliable activations.

\textbf{Cross-subject generalization:} Domain adaptation techniques including adversarial alignment \citep{ganin2016dann}, distribution matching, and adaptive transfer learning \citep{zhang2021adaptive} have been studied to reduce calibration requirements. MultiDiffNet \citep{zhang2025multidiffnet} takes a different route, learning a shared latent space jointly optimized via diffusion, contrastive, and reconstruction objectives to improve cross-subject generalization without explicit distribution alignment.

\section{Methodology}

\subsection{Datasets}
We evaluated our framework across 3 EEG paradigms (SSVEP, P300, and Motor Imagery) using 6 benchmark datasets: Wang2016 \citep{wang2016} and Lee2019 \citep{lee2019} for SSVEP; BI2014b \citep{bi2014b} and BNCI2014\_009 \citep{bnci2014009} for P300; and BNCI2014\_001 \citep{bnci2014001} and Lee2019 \citep{lee2019} for MI. Data were partitioned into training, validation, and two distinct test sets: a seen-subject (cross-session) split and an unseen-subject (cross-subject) split to evaluate generalization. Dataset specifications are summarized in Appendix \ref{apdx: dataset}.

\subsection{Domain Characterization}

\begin{figure} 
    \centering
    \includegraphics[width=1.0\linewidth]{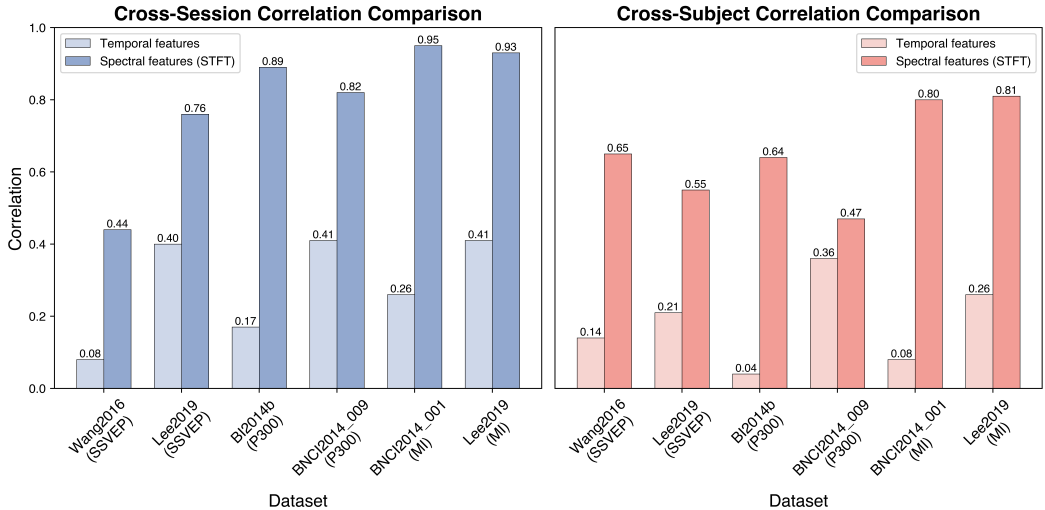}
    \caption{Cross-session (left) and cross-subject (right) correlation comparison between temporal and spectral representations. Spectral features exhibit consistently higher cross-subject similarity across all datasets.}
    \label{fig:tempvsspec}
    \vspace{-10pt}
\end{figure}

To inform our architectural choices, we carried out systematic correlation analyses in both the temporal and spectral domains across all EEG datasets. For each dataset, we constructed class-conditional representative patterns by averaging all trials sharing the same label, producing a time-domain representation of shape $(C, T)$. For the spectral analysis, we applied an STFT to each trial (using task-specific parameters from the corresponding dataset configuration), took the power representation $|Z(f,t)|^2$, and averaged within each class to obtain a spectral representative pattern of shape $(C, F, T')$. To quantify the similarity between any two representative patterns, we flattened and z-normalized each one, then computed their Pearson correlation coefficient.

We examined two complementary notions of signal consistency: cross-session and cross-subject correlation. Cross-session correlation measures the agreement between representative patterns derived from different recording sessions of the same subject, aggregated across all session pairs and labels. Cross-subject correlation measures the agreement between representative patterns from different subjects, again aggregated across all subject pairs and labels. As shown in Figure~\ref{fig:tempvsspec}, spectral representative patterns consistently exhibited higher cross-subject similarity than their raw temporal counterparts across datasets. This suggests that time-frequency representations capture a more subject-invariant signature of task-relevant neural dynamics, and directly motivates our modeling framework's emphasis on spectral representations as a robust foundation for cross-subject generalization.

\subsection{Signal Processing}

We process the raw EEG signals into two modalities to accommodate the dual-stream architecture of ASPEN.

\textbf{Temporal Modality:} To ensure signal stability and cross-subject consistency, raw EEG data are preprocessed using task-specific bandpass filtering and trial-wise Z-score normalization to achieve zero mean and unit variance. Frequency ranges are tailored to each paradigm: 6-90 Hz for SSVEP to capture high-frequency harmonics \citep{wang2016}, 1-24 Hz for P300 to isolate the evoked response, and 4-40 Hz for Motor Imagery (MI) to focus on $\mu$ and $\beta$ rhythms \citep{lee2019}. Signals are downsampled to optimize computational efficiency (see Appendix \ref{apdx: dataset} for details).

\textbf{Spectral Modality:} To obtain frequency-domain representations, we apply the Short-Time Fourier Transform (STFT) to each preprocessed trial on a per-channel basis. For each paradigm, we fix a task-specific set of STFT parameters $(f_s, n_{\mathrm{perseg}}, n_{\mathrm{overlap}}, n_{\mathrm{fft}})$ drawn from the dataset configuration. These parameters are validated through a 27-way ablation study with varying window length, overlap ratio, and Fast Fourier Transform (FFT) size, ensuring that all subjects share a consistent $\mathrm{frequency}\times \mathrm{time}$ grid at the input of the spectral stream. Concretely, we compute spectrograms with a Hann window and convert them to power spectra by taking the squared magnitude $|Z_c(f,t)|^2$ for each channel $c$, yielding a 3D tensor of shape $(C,F,T)$ per trial. For the purposes of our comparative analysis, we refer to the standalone Spectral Encoder Network as SPEN, while ASPEN represents the full hybrid framework utilizing multiplicative fusion. These power spectrograms are fed into the SPEN convolutional blocks, where the chosen window and hop sizes control the trade-off between temporal resolution $\Delta t=(n_{\mathrm{perseg}}-n_{\mathrm{overlap}})/f_s$ and frequency resolution $\Delta f=f_s/n_{\mathrm{fft}}$, allowing ASPEN to capture task-relevant harmonic structure for SSVEP, evoked components for P300, and $\mu/\beta$ band dynamics for MI in a unified spectral representation.

\subsection{Model Architecture}

\begin{figure*} [h]	
    \centering
    \includegraphics[width=1\linewidth]{fig3.png}
    \caption{Detailed view of temporal stream, spectral stream, and multiplicative fusion components.}
    \label{fig:model_details}
    \vspace{-10pt}
\end{figure*}

The high-level architecture is shown in Figure~\ref{fig:model}. ASPEN consists of two primary components: a Temporal Stream and a Spectral Stream. Given a raw EEG trial $\mathbf{X} \in \mathbb{R}^{C \times T}$, the two complementary inputs are the raw signal $\mathbf{X}_{\text{time}}$ for the temporal stream and the per-channel STFT magnitude spectrograms $\mathbf{X}_{\text{spec}} \in \mathbb{R}^{C \times F \times T'}$ for the spectral stream.

A detailed view of the two streams and the fusion mechanism are illustrated in Figure\ref{fig:model_details}. The spectral stream extracts frequency-time patterns through a two-stage CNN with squeeze-and-excitation (SE) attention~\citep{hu2018squeeze} and residual blocks. SE modules adaptively recalibrate channel responses by learning to emphasize informative spectral patterns, while residual connections improve gradient flow. After two stages of convolution, SE attention, and pooling, features are projected and averaged across EEG channels to yield $\mathbf{x}_s \in \mathbb{R}^{d}$. The temporal stream follows an EEGNet-inspired design~\citep{lawhern2018eegnet} where the temporal convolution learns frequency-specific filters analogous to bandpass filtering, the depthwise spatial convolution learns channel combinations analogous to CSP~\citep{ang2008fbcsp}, and the separable convolutions efficiently extract higher-order features. The output is projected to $\mathbf{x}_t \in \mathbb{R}^{d}$.

We combine stream representations via element-wise multiplication after learned linear projections
\begin{equation}
    \mathbf{z} = (\mathbf{W}_s \mathbf{x}_s) \odot (\mathbf{W}_t \mathbf{x}_t)
    \label{eq:fusion}
\end{equation}
where $\mathbf{W}_s, \mathbf{W}_t \in \mathbb{R}^{d \times d}$ are learnable matrices and $\odot$ denotes the Hadamard product. This multiplicative interaction acts as cross-modal gating. Dimension $z_i$ is large only when both streams produce strong activations, effectively requiring agreement between spectral and temporal evidence. Features that appear prominently in only one view, which often indicative of artifacts or noise, are suppressed. The fused representation passes through batch normalization and a linear classifier.

The model is optimized using task-specific loss functions. Binary cross-entropy with logits ($\mathcal{L}_{BCE}$) are used for two-class paradigms like P300, which includes automated positive-weight scaling to mitigate class imbalance. For multi-class tasks such as SSVEP and Motor Imagery, standard cross-entropy loss ($\mathcal{L}_{CE}$) is employed. The learned weights $w_s$ and $w_t$ provide an interpretable measure of each modality's contribution, enabling an analysis of which representation the model prioritizes for different paradigms and individual trials.

\section{Experiments}

\subsection{Baselines}
To evaluate the performance of our proposed method, we benchmarked against five baselines that emphasize cross-subject generalization and novel data representations.

EEGNet \citep{lee2019} serves as a compact convolutional baseline, leveraging depthwise and separable convolutions to efficiently extract spatial and frequency-specific features with minimal parameters. To model global dependencies, EEGConformer \citep{song2022eeg} adopts a hybrid design that combines CNNs for local feature extraction with Transformer modules for long-range temporal modeling. CTNet \citep{zhao2024ctnet} is included for its emphasis on cross-task and cross-subject robustness, utilizing domain-invariant representations to mitigate EEG non-stationarity. TSformer-SA \citep{li2025tsformer} integrates temporal and spectral features through cross-view self-attention, enabling joint modeling of time-domain signals and wavelet-based time-frequency representations for improved cross-subject decoding. Finally, MultiDiffNet \citep{zhang2025multidiffnet} incorporates multi-scale differential transformations of the input signal to better capture complex distributions and enhance training stability in noisy data environments.

All models were evaluated using identical training schedules and hyperparameters unless otherwise specified by architectural constraints. The performance of these baselines are shown in Table \ref{tab:main_results}.

\subsection{Ablations}
\begin{table*}[htbp]
\centering
\caption{Ablation study summary. Best STFT parameters and fusion strategy per dataset, selected by unseen-subject accuracy. Best Acc = best fusion accuracy on held-out subjects (\%), Mult Acc = multiplicative fusion accuracy (\%), $\Delta$ = absolute difference. Global Attn = Global Attention, Bilinear = Low-rank Bilinear.}
\label{tab:fusion_ablation_summary}
\vspace{1em}  
\small
\begin{tabular}{l|ccc|cccc}
\toprule
\textbf{Dataset} & \textbf{nperseg} & \textbf{noverlap} & \textbf{nfft} & \textbf{Best Fusion} & \textbf{Best Acc} & \textbf{Mult Acc} & \textbf{$\Delta$} \\
\midrule
Wang 2016 SSVEP   & 256 & 128 & 256  & Global Attn    & 72.76 & 69.47 & -3.3 \\
Lee2019 SSVEP     & 256 & 128 & 1024 & Bilinear       & 86.68 & 85.71 & -1.0 \\
BI2014b P300      & 32  & 16  & 512  & Bilinear       & 73.52 & 66.12 & -7.4 \\
BNCI2014-009 P300 & 128 & 120 & 256  & Multiplicative & 89.82 & 89.82 & -- \\
BNCI2014-001 MI   & 512 & 256 & 512  & Multiplicative & 30.73 & 30.73 & -- \\
Lee2019 MI        & 32  & 30  & 32   & Multiplicative & 75.70 & 75.70 & -- \\
\bottomrule
\end{tabular}
\end{table*}

\textbf{STFT Settings:} Since SPEN operates on time-frequency representations, we first optimized the STFT frontend through a controlled ablation study. For each task, we evaluated 27 configurations by sweeping three values per parameter: window length (\texttt{nperseg}), overlap ratio (0.50, 0.75, 0.9375), and FFT size (\texttt{nfft}). Window lengths were defined in a task-aware manner (half, default, and maximum resolution), constrained to never exceed trial length; \texttt{noverlap} was derived from the overlap ratio subject to \texttt{noverlap} $<$ \texttt{nperseg}; and \texttt{nfft} was drawn from a pool containing the smallest power of two $\geq$ \texttt{nperseg} and the task default, with \texttt{nfft} $\geq$ \texttt{nperseg} enforced.

For each configuration, we trained the same SPEN backbone and training protocol to isolate the effect of STFT settings, then evaluated on validation and test splits using accuracy and F1/recall (plus ROC-AUC and PR-AUC for imbalanced binary P300 tasks). For binary tasks, we re-optimized the decision threshold on the validation set (F1-maximizing sweep) and applied it to test evaluation; for P300 tasks we additionally used a WeightedRandomSampler to balance training batches. The best STFT setting was selected per task (F1 for P300, accuracy otherwise), and the top 3 STFT configurations were carried forward into subsequent fusion ablations to avoid confounding fusion comparisons with suboptimal preprocessing. Full details are provided in Appendix \ref{apdx: stft}.

\textbf{Fusion Strategies:} We evaluated seven fusion strategies for combining the temporal and spectral streams, drawing from foundational methods in multimodal learning \citep{liang2023foundationstrendsmultimodalmachine}: (1) static equal weighting, (2) global attention with learned trial-level weights, (3) spatial attention with per-channel weighting, (4) gated linear units (GLU) for noise suppression, (5) element-wise multiplicative fusion, (6) low-rank bilinear pooling, and (7) multi-head cross-attention between streams. Each strategy was evaluated across all benchmark tasks using the top 3 best performing STFT parameters from our spectrogram ablation.

While optimal fusion varied by dataset, multiplicative fusion achieved the highest unseen-subject accuracy on three of six tasks (BNCI2014 P300, BNCI2014-001 MI, Lee2019 MI) and remained competitive on two others (within 1\% on Wang2016 SSVEP and 3.3\% on Lee2019 SSVEP). Bilinear fusion outperforms on BI2014b P300 by 7.4\%, but we prioritize cross-paradigm consistency over peak single-task performance. Given multiplicative fusion's stable cross-paradigm performance and its alignment with our cross-modal gating hypothesis (Equation \ref{eq:fusion}), we adopt it as the unified fusion strategy for ASPEN. This selection also determined which STFT configuration to use for final evaluation. Best fusion method and STFT parameters are summarized in Table \ref{tab:fusion_ablation_summary}. Full mathematical details and per-task results are provided in Appendix \ref{apdx: fusion}.

\subsection{Results}

\begin{table*}[htbp]
\centering
\caption{Final results across tasks and models. Cross-subject generalization accuracy (\%). Bold indicates best cross-subject performance per dataset. Mean ± STD across three seeds.}
\label{tab:main_results}
\tiny 
\begin{tabular}{lll ccccccc}
\toprule
& & & \multicolumn{7}{c}{\scriptsize \textbf{Method}} \\ 
\cmidrule(lr){4-10}
{\scriptsize \textbf{Task}} & {\scriptsize \textbf{Dataset}} & {\scriptsize \scriptsize{Cross-}} & {EEGNet} & {EEGConf.} & {MultiDiff.} & {TSformer-SA} & {CTNet} & {SPEN} & {ASPEN} \\
\midrule
\multirow{4}{*}{\scriptsize \textbf{SSVEP}} 
& \multirow{2}{*}{Wang2016} 
& Session   & 81.96$\pm$6.02 & 56.32$\pm$4.15 & 91.74$\pm$1.62 & 47.96$\pm$4.16 & 88.37$\pm$4.20 & 85.71$\pm$1.57 & 73.98$\pm$1.46 \\
& & Subject & 74.25$\pm$5.27 & 49.95$\pm$2.81 & \textbf{87.95}$\pm$2.56 & 39.93$\pm$5.25 & 83.60$\pm$0.82 & 78.82$\pm$6.41 & 67.20$\pm$4.83 \\
\cmidrule(lr){2-10}
& \multirow{2}{*}{Lee2019} 
& Session   & 95.04$\pm$0.35 & 92.00$\pm$0.94 & 93.67$\pm$0.67 & 79.98$\pm$6.60 & 95.36$\pm$0.61 & 70.58$\pm$0.80 & 95.50$\pm$0.33 \\
& & Subject & 86.51$\pm$0.09 & 81.99$\pm$1.05 & 85.04$\pm$0.76 & 63.38$\pm$6.43 & 87.25$\pm$0.69 & 58.51$\pm$0.98 & \textbf{87.53}$\pm$0.29 \\
\midrule
\multirow{4}{*}{\scriptsize \textbf{P300}} 
& \multirow{2}{*}{BI2014b} 
& Session   & 64.74$\pm$2.71 & 78.84$\pm$3.46 & 81.46$\pm$4.56 & 82.25$\pm$3.04 & 76.06$\pm$6.45 & 84.15$\pm$0.93 & 77.95$\pm$5.52 \\
& & Subject & 62.64$\pm$1.01 & 77.55$\pm$4.59 & 80.95$\pm$3.91 & \textbf{83.13}$\pm$0.35 & 74.55$\pm$8.41 & 82.96$\pm$0.64 & 77.01$\pm$7.16 \\
\cmidrule(lr){2-10}
& \multirow{2}{*}{BNCI2014} 
& Session   & 84.65$\pm$0.62 & 84.25$\pm$1.09 & 85.42$\pm$0.89 & 86.75$\pm$0.68 & 86.26$\pm$0.38 & 78.31$\pm$4.40 & 89.65$\pm$0.48 \\
& & Subject & 84.05$\pm$1.84 & 83.20$\pm$1.29 & 84.91$\pm$1.68 & 86.92$\pm$1.06 & 85.28$\pm$0.90 & 77.97$\pm$4.66 & \textbf{88.57}$\pm$0.76 \\
\midrule
\multirow{4}{*}{\scriptsize \textbf{MI}} 
& \multirow{2}{*}{BNCI2014} 
& Session   & 61.21$\pm$3.27 & 56.35$\pm$3.05 & 58.53$\pm$4.46 & 35.91$\pm$6.48 & 57.34$\pm$7.68 & 33.63$\pm$3.32 & 51.59$\pm$8.40 \\
& & Subject & \textbf{37.29}$\pm$4.45 & 33.91$\pm$7.33 & 29.05$\pm$4.00 & 28.99$\pm$4.82 & 36.29$\pm$9.82 & 26.91$\pm$3.69 & 32.00$\pm$0.53 \\
\cmidrule(lr){2-10}
& \multirow{2}{*}{Lee2019} 
& Session   & 78.89$\pm$0.38 & 76.28$\pm$0.91 & 76.54$\pm$2.47 & 71.90$\pm$0.74 & 77.15$\pm$0.77 & 55.28$\pm$0.45 & 77.93$\pm$0.52 \\
& & Subject & 75.88$\pm$1.98 & 74.55$\pm$1.39 & 74.67$\pm$1.95 & 71.40$\pm$2.61 & 74.98$\pm$2.13 & 53.50$\pm$1.69 & \textbf{76.27}$\pm$0.73 \\
\bottomrule
\end{tabular}
\end{table*}

The performance of SPEN, ASPEN, and five baselines across six benchmark datasets is summarized in Table~\ref{tab:main_results}. Our results demonstrate that ASPEN achieves superior cross-subject generalization in three of the six evaluated datasets: Lee2019 SSVEP (87.53\%), BNCI2014 P300 (88.57\%), and Lee2019 MI (76.27\%) datasets. Notably, on the BNCI2014 P300 task, ASPEN outperforms TSformer-SA by nearly 2\%, despite the latter being specifically designed for evoked potential decoding. This suggests that our multiplicative gating mechanism is more effective at filtering the inter-subject noise prevalent in large-scale P300 datasets.

We also observe that while TSformer-SA performs competitively on P300-like tasks (83.13\% on BI2014b), its performance degrades significantly on SSVEP and MI tasks (39.93\% on Wang2016 SSVEP). In contrast, ASPEN maintains competitive performance across all three tasks. This indicates that multiplicative fusion acts as a universal architectural prior that adapts to the specific spectral-temporal demands of the underlying neural signal.

The standalone spectral encoder (SPEN) struggles on Motor Imagery tasks, achieving only 26.91\% on BNCI2014-001 MI (barely above the 25\% chance level) and 53.50\% on Lee2019 MI. These results reveal that cross-subject stability and discriminative power are distinct properties. While spectral representations are more consistent across individuals, Motor Imagery classification relies on precise temporal dynamics of sensorimotor rhythms that are lost in the STFT magnitude representation. The performance recovery from SPEN to ASPEN on Lee2019 MI (53.50\% to 76.27\%) demonstrates that neither modality alone suffices and that cross-modal fusion is essential for robust generalization.

\section{Analysis}
\label{sec:analysis}

Our experimental results demonstrate that ASPEN achieves superior or competitive cross-subject generalization compared to state-of-the-art baselines. To understand the drivers of this performance, we analyze the impact of multiplicative fusion, the paradigm-specific reliance on spectral versus temporal features, and the role of spectral stability.

\subsection{Mechanism of Multiplicative Spectral-Temporal Gating}

\begin{figure*}[t] 
    \centering
    \includegraphics[width=0.7\linewidth]{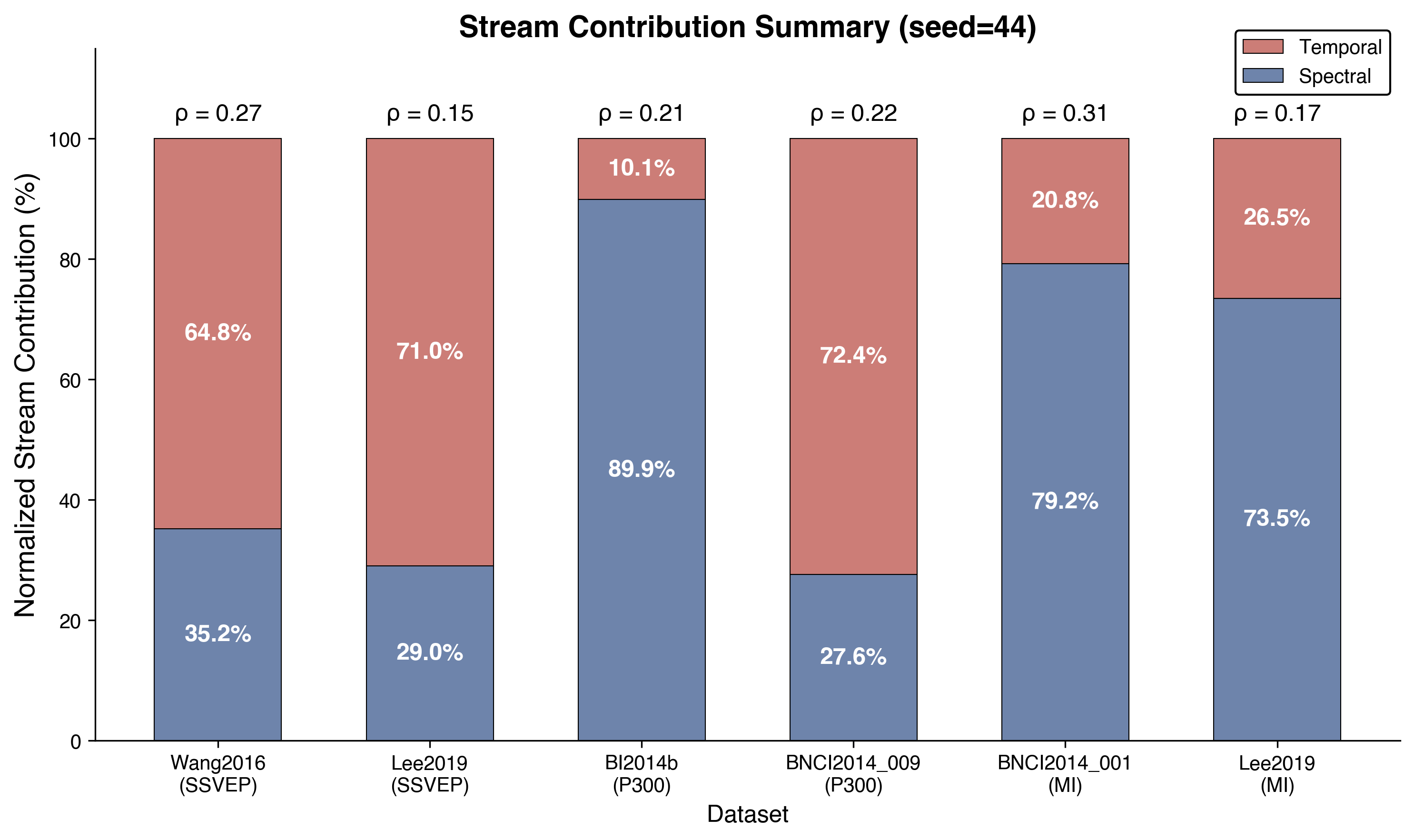}
    \caption{Stream contributions and feature correlation ($\rho$) across datasets. Low correlation values confirm that streams capture distinct information.}
    \label{fig:weighting}
    \vspace{-10pt}
\end{figure*}

To investigate how ASPEN leverages dual-stream information, we analyze the features during inference. The fused representation is defined as:
\begin{equation}
    \mathbf{z}_{\text{fused}} = \text{proj}_S(\mathbf{x}_S) \odot \text{proj}_T(\mathbf{x}_T)
\end{equation}
where $\mathbf{x}_S$ and $\mathbf{x}_T$ denote the spectral and temporal features respectively, and $\odot$ represents element-wise multiplication. We quantify the relative spectral magnitude $w_S$ via the normalized L2 norm of the projected features:
\begin{equation}
    w_S = \frac{\|\text{proj}_S(\mathbf{x}_S)\|_2}{\|\text{proj}_S(\mathbf{x}_S)\|_2 + \|\text{proj}_T(\mathbf{x}_T)\|_2}
\end{equation}
with $w_T = 1 - w_S$ representing the relative temporal magnitude. Stream complementarity is measured through the feature correlation $\rho$, defined as the cosine similarity between the projected features:
\begin{equation}
    \rho = \frac{\langle \text{proj}_S(\mathbf{x}_S), \text{proj}_T(\mathbf{x}_T) \rangle}{\|\text{proj}_S(\mathbf{x}_S)\|_2 \cdot \|\text{proj}_T(\mathbf{x}_T)\|_2}
\end{equation}

As illustrated in Figure~\ref{fig:weighting}, ASPEN adaptively shifts its reliance on spectral vs. temporal features based on the task at hand. The P300 task (BI2014b) exhibits strong spectral dominance ($w_S = 89.9\%$), while SSVEP tasks (Wang2016, Lee2019) shift toward the temporal stream, with $w_T$ reaching $64.8\%$ and $71.0\%$, respectively. Motor imagery datasets show a spectrally-biased distribution ($w_S \approx 73\%$--$79\%$). Across all datasets, the low correlation values ($0.15 \leq \rho \leq 0.31$) confirm that the streams capture distinct, non-redundant information.

This architecture functions as a strict cross-modal gating mechanism. Unlike additive fusion, where high-magnitude artifacts in one modality can bias the decision boundary, our multiplicative approach acts as a logical \textit{AND} gate. A feature is only activated in the fused representation if it receives concurrent support from both streams. Consequently, transient artifacts, such as muscle noise that appears in the temporal domain but lacks spectral consistency, are naturally suppressed. As evidenced in Table~\ref{tab:fusion_ablation_summary}, forcing this cross-view agreement encourages the model to prioritize features robust to the phase shifts and amplitude variations inherent in cross-subject transfer.

\subsection{Visualizing Decision Boundaries}
\begin{figure}
    \centering
    \includegraphics[width=1\linewidth]{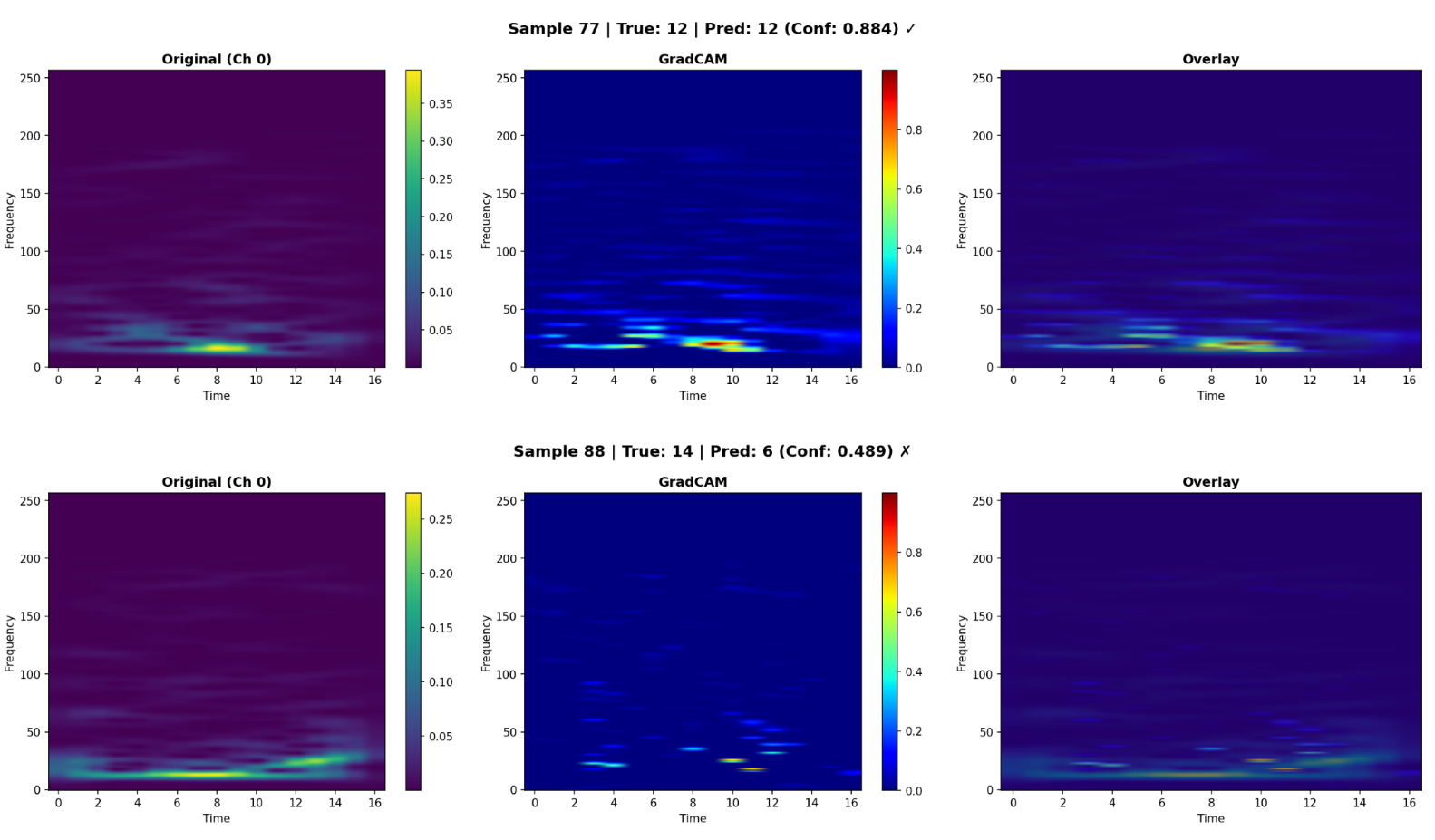}
    \caption{Grad-CAM visualization of feature importance for P300 classification. Correct prediction (top) shows focused attention on physiologically relevant low-frequency bands. Misclassification (bottom) reveals scattered attention towards high-frequency noise artifacts.}
    \label{fig:gradcam}
    \vspace{-10pt}
\end{figure} 

To validate the interpretability of our framework and justify the necessity of cross-modal fusion, we visualized the learned features using Grad-CAM \cite{selvaraju2017grad}. We analyzed the spectral regions contributing most to the model's decisions in both successful and failed prediction scenarios on the P300 dataset.

Fig.\ref{fig:gradcam} illustrates the Grad-CAM activation maps for two representative samples. As shown in the top row of Fig.~\ref{fig:gradcam}, when the model \textit{correctly} identifies the target class with high confidence, the activation hotspot is tightly concentrated in the low-frequency band and specific temporal windows. This aligns perfectly with neurophysiological knowledge, as P300 components are primarily characterized by low-frequency energy deflections. The model successfully ignores high-frequency background activity, confirming that it has learned robust, physiologically valid features.

Conversely, the bottom row of Fig.~\ref{fig:gradcam} shows a misclassified sample with low confidence. Here, the model's attention is fragmented and scattered across high-frequency bands, likely driven by muscle artifacts or instrument noise rather than neural signals. This distraction by high-frequency noise highlights the vulnerability of single-stream interactions where artifactual high-amplitude spikes can propagate to the decision layer.

\section{Conclusion and Outlook}

In this work, we introduced ASPEN, a multimodal framework designed to overcome the challenges of cross-subject generalization in EEG-based BCIs. By leveraging the inherent stability of spectral representations, ASPEN utilizes a multiplicative fusion mechanism to enforce cross-modal agreement. Our experiments across six benchmark datasets show that this approach effectively suppresses non-neural artifacts and prioritizes robust features, achieving the best unseen-subject performance on three datasets spanning SSVEP, P300, and Motor Imagery paradigms.

While ASPEN significantly reduces the performance gap for new users, several avenues for future research remain. Future work will investigate automated configuration optimization, perhaps through learnable time-frequency transforms, to move toward a truly "one-size-fits-all" zero-shot model. Additionally, we aim to explore the integration of self-supervised pre-training on large multi-subject EEG corpora to further enhance the richness of the shared latent space.


\subsubsection*{Acknowledgments}
We would like to thank Professor Bhiksha Raj of Carnegie Mellon University for his guidance and support throughout this project. This work was partly supported by Institute of Information \& communications Technology Planning \& Evaluation (IITP) grant funded by the Korea government(MSIT) (RS-2022-00143911, AI Excellence Global Innovative Leader Education Program).

\bibliography{main}
\bibliographystyle{iclr2026_conference}

\appendix
\section*{Appendix}
\renewcommand{\thesubsection}{\Alph{subsection}}

\subsection{Dataset Specifications and Splitting}
\label{apdx: dataset}

To evaluate model robustness across datasets, we employed a multi-tier splitting strategy with a 60/20/20 ratio for subjects included in training. For each task, a subset of subjects was designated as ``seen'' and their data was partitioned into three splits. The Training split (60\%) was used for model optimization, the Validation split (20\%) for hyperparameter tuning, and Test 1 (20\%) for cross-session evaluation. Test 1 specifically evaluates the model's ability to generalize across different recording sessions from the same individuals, thereby assessing robustness to within-subject temporal variations.

To assess zero-shot generalization capabilities, a separate cohort of subjects was entirely withheld from the training process. Test 2 (Cross-Subject) evaluates model performance on individuals with previously unencountered physiological profiles, providing a rigorous test of subject-independent performance. 

\begin{table}[h]
  \caption{EEG Dataset Specifications and Preprocessing Parameters. Z-score normalization was applied to all datasets to standardize signal amplitudes across different subjects and sessions, ensuring that high-voltage artifacts or individual physiological variations do not disproportionately influence model training.}
  \label{tab:datasets}
  \begin{center}
    \small
    \setlength{\tabcolsep}{4pt}
    \begin{tabular}{llcccccc}
      \hline \\
      \bf Task & \bf Dataset & \bf Subj. & \bf Chan. & \bf Cls. & \bf Bandpass & \bf Epoch & \bf SR \\ 
      \hline \\
      \textbf{SSVEP}: Frequency- & Wang2016 & 35 & 64 & 26 & 6--90 Hz & 1.0s & 250 Hz \\
      tagged visual decoding & Lee2019 & 54 & 62 & 4 & 6--90 Hz & 1.0s & 250 Hz \\ 
      \hline \\
      \textbf{P300}: Binary target & BI2014b & 38 & 32 & 2 & 0.1--30 Hz & 1.0s & 256 Hz \\
      vs. non-target ERP & BNCI2014\_009 & 10 & 16 & 2 & 1--24 Hz & 1.0s & 256 Hz \\ 
      \hline \\
      \textbf{MI}: Motor & BNCI2014\_001 & 9 & 22 & 4 & 4--40 Hz & 4.0s & 250 Hz \\
      imagery decoding & Lee2019 & 54 & 22 & 2 & 4--40 Hz & 4.0s & 250 Hz \\ 
      \hline
    \end{tabular}
  \end{center}
\end{table}

\subsection{STFT Ablation}
\subsubsection{STFT Methods}
\label{apdx: stft}

\label{app:stft}

\paragraph{STFT-based Spectral Representation.}
For each EEG trial, we construct a spectral (time--frequency) representation by applying the Short-Time Fourier Transform (STFT) independently to each channel. Let $x_c[n]$ denote the preprocessed discrete-time signal of channel $c\in\{1,\dots,C\}$ with sampling rate $f_s$ (Hz), and let $w[m]$ be a Hann window of length $n_{\mathrm{perseg}}$. We compute the complex STFT as
\begin{equation}
Z_c(f_k,t_\ell)
= \sum_{m=0}^{n_{\mathrm{perseg}}-1}
x_c[\ell H + m]\; w[m]\; e^{-j 2\pi k m / n_{\mathrm{fft}}},
\label{eq:stft}
\end{equation}
where $n_{\mathrm{fft}}$ is the FFT size, $H=n_{\mathrm{perseg}}-n_{\mathrm{overlap}}$ is the hop size (in samples), and $(f_k,t_\ell)$ index frequency and time frames, respectively. In practice, we use the one-sided spectrum for real-valued signals, yielding $F = \lfloor n_{\mathrm{fft}}/2\rfloor + 1$ frequency bins. We then convert complex spectrograms to \emph{power} spectrograms by taking the squared magnitude:
\begin{equation}
S_c(f_k,t_\ell) = |Z_c(f_k,t_\ell)|^2.
\label{eq:power}
\end{equation}
Each trial is thus represented as a tensor $\mathbf{S}\in\mathbb{R}^{C\times F\times T}$, where $T$ denotes the number of time frames determined by the trial length and hop size. These power spectrograms are provided directly to the spectral stream of the model, ensuring a consistent $(F\times T)$ grid across subjects within each paradigm. The STFT parameters control the time--frequency trade-off: the frequency resolution is $\Delta f = {f_s}/{n_{\mathrm{fft}}}$ (Hz per bin) and the frame step is $\Delta t = {H}/{f_s} = {(n_{\mathrm{perseg}}-n_{\mathrm{overlap}})}/{f_s}$ seconds. Larger $n_{\mathrm{perseg}}$ and $n_{\mathrm{fft}}$ improve frequency resolution, while higher overlap (smaller $H$) improves temporal granularity.

\paragraph{Task-aware STFT parameter search space.}
Because optimal STFT settings can vary by paradigm and trial length, we perform a systematic ablation over STFT parameters for each task. The ablation evaluates a fixed set of 27 configurations formed by selecting three values for each of: (i) window length $n_{\mathrm{perseg}}$, (ii) overlap ratio $r\in\{0.50,\,0.75,\,0.9375\}$ with $n_{\mathrm{overlap}}=\lfloor r\,n_{\mathrm{perseg}}\rfloor$ and the hard constraint $n_{\mathrm{overlap}}<n_{\mathrm{perseg}}$, and (iii) FFT size $n_{\mathrm{fft}}$. To ensure feasibility, we enforce $n_{\mathrm{perseg}}\le L_{\mathrm{trial}}$ where $L_{\mathrm{trial}}$ is the number of samples per trial for the dataset. Window-length candidates are generated from the task default $n_{\mathrm{perseg}}^{(0)}$ as
\begin{equation}
\mathcal{N}_{\mathrm{perseg}} =
\Big\{\max(32,\lfloor \tfrac{1}{2}n_{\mathrm{perseg}}^{(0)}\rfloor),\;
n_{\mathrm{perseg}}^{(0)},\;
\min(L_{\mathrm{trial}},\,2n_{\mathrm{perseg}}^{(0)})\Big\},
\end{equation}
after removing duplicates. For each $n_{\mathrm{perseg}}$, we build an FFT pool
\begin{equation}
\mathcal{P}=\mathrm{unique}\Big(\{2^{\lceil \log_2(n_{\mathrm{perseg}})\rceil},\,256,\,512,\,1024,\,2048,\,n_{\mathrm{fft}}^{(0)}\}\Big),
\end{equation}
retain only valid $n_{\mathrm{fft}}\ge n_{\mathrm{perseg}}$, and choose three FFT candidates centered around $n_{\mathrm{fft}}^{(0)}$ when available (otherwise the first three valid values). If fewer than three candidates remain, we pad by doubling the maximum value until three candidates are obtained. Each configuration is assigned an identifier
\begin{equation}
\texttt{nperseg}\{n_{\mathrm{perseg}}\}\_\texttt{ov}\{100r\}\_\texttt{nfft}\{n_{\mathrm{fft}}\}.
\end{equation}

\paragraph{Ablation protocol and model selection.}
For each task and STFT configuration, we train the same model architecture with identical optimization hyperparameters to isolate the effect of the STFT frontend. We report accuracy and complementary metrics (macro-F1/macro-recall for multi-class; precision/recall, ROC-AUC, and PR-AUC for binary tasks). For binary tasks, we additionally optimize the decision threshold on the validation set by sweeping $t\in\{0.01,0.02,\dots,0.99\}$ and maximizing validation F1, then reuse the selected threshold for all test evaluations to avoid test-time tuning. For P300-style binary datasets, we mitigate class imbalance using weighted sampling, assigning example weights inversely proportional to class frequency. The best STFT configuration is selected per task using validation accuracy for multi-class tasks and validation PR-AUC (fallback to ROC-AUC if needed) for binary P300 tasks; we log all configurations and metrics (CSV/JSON) to enable reproducibility.

\subsubsection{STFT Ablation Results}

\begin{table}[H]
\centering
\caption{STFT Ablation SSVEP}
\label{tab:ssvep_results}
\footnotesize
\begin{tabular}{ccc|c|cc|ccccc}
\toprule
 & & & val & seen & seen & unseen & unseen & unseen & unseen & unseen \\
nperseg & noverlap & nfft & acc & acc & loss & acc & f1 & recall & auc & loss \\
\midrule
\textbf{128} & \textbf{64} & \textbf{256} & \textbf{83.05} & \textbf{80.06} & \textbf{0.77} & \textbf{71.71} & \textbf{71.90} & \textbf{71.71} & \textbf{96.08} & \textbf{1.15} \\
128 & 64 & 512 & 82.34 & 81.20 & 0.69 & 69.23 & 69.25 & 69.23 & 96.16 & 1.17 \\
128 & 64 & 1024 & 81.34 & 80.06 & 0.76 & 65.46 & 65.71 & 65.46 & 95.30 & 1.26 \\
128 & 96 & 256 & 83.76 & 82.76 & 0.70 & 70.83 & 70.91 & 70.83 & 96.21 & 1.13 \\
128 & 96 & 512 & 82.48 & 82.91 & 0.66 & 71.31 & 71.41 & 71.31 & 96.57 & 1.10 \\
128 & 96 & 1024 & 82.19 & 83.33 & 0.66 & 70.11 & 70.18 & 70.11 & 96.06 & 1.13 \\
128 & 120 & 256 & 82.19 & 83.05 & 0.69 & 70.27 & 70.34 & 70.27 & 96.09 & 1.14 \\
128 & 120 & 512 & 81.77 & 81.77 & 0.70 & 70.35 & 70.57 & 70.35 & 96.02 & 1.13 \\
\textbf{256} & \textbf{128} & \textbf{256} & \textbf{84.47} & \textbf{82.91} & \textbf{0.69} & \textbf{72.60} & \textbf{72.69} & \textbf{72.60} & \textbf{96.75} & \textbf{1.08} \\
\textbf{256} & \textbf{128} & \textbf{512} & \textbf{83.76} & \textbf{84.19} & \textbf{0.64} & \textbf{71.47} & \textbf{71.59} & \textbf{71.47} & \textbf{96.82} & \textbf{1.07} \\
256 & 128 & 1024 & 84.19 & 83.33 & 0.64 & 68.99 & 69.30 & 68.99 & 95.83 & 1.19 \\
256 & 192 & 256 & 81.91 & 82.19 & 0.72 & 71.39 & 71.54 & 71.39 & 96.05 & 1.13 \\
256 & 192 & 512 & 82.48 & 82.62 & 0.67 & 71.31 & 71.51 & 71.31 & 95.98 & 1.15 \\
256 & 192 & 1024 & 83.19 & 82.34 & 0.67 & 70.59 & 70.80 & 70.59 & 96.06 & 1.16 \\
256 & 240 & 256 & 83.62 & 84.47 & 0.64 & 70.35 & 70.48 & 70.35 & 96.23 & 1.14 \\
256 & 240 & 512 & 83.33 & 83.48 & 0.65 & 69.55 & 69.62 & 69.55 & 95.91 & 1.15 \\
\bottomrule
\end{tabular}
\end{table}

\begin{table}[H]
\centering
\caption{STFT Ablation Lee2019 SSVEP}
\label{tab:lee2019_ssvep_results}
\footnotesize
\begin{tabular}{ccc|c|cc|ccccc}
\toprule
 & & & val & seen & seen & unseen & unseen & unseen & unseen & unseen \\
nperseg & noverlap & nfft & acc & acc & loss & acc & f1 & recall & auc & loss \\
\midrule
64 & 32 & 256 & 61.06 & 61.09 & 0.95 & 48.14 & 48.22 & 48.14 & 72.63 & 1.28 \\
64 & 32 & 512 & 67.09 & 67.00 & 0.85 & 53.70 & 53.74 & 53.70 & 77.73 & 1.18 \\
64 & 32 & 1024 & 66.19 & 65.75 & 0.87 & 53.12 & 53.20 & 53.12 & 77.35 & 1.14 \\
64 & 48 & 256 & 69.34 & 68.22 & 0.85 & 54.77 & 54.77 & 54.77 & 78.36 & 1.21 \\
64 & 48 & 512 & 69.28 & 68.13 & 0.81 & 55.52 & 55.56 & 55.52 & 78.56 & 1.14 \\
64 & 48 & 1024 & 66.63 & 66.09 & 0.87 & 53.50 & 53.53 & 53.50 & 77.09 & 1.22 \\
64 & 60 & 256 & 58.66 & 60.00 & 0.98 & 46.52 & 46.39 & 46.52 & 71.48 & 1.24 \\
64 & 60 & 512 & 67.31 & 66.44 & 0.85 & 52.84 & 52.91 & 52.84 & 76.35 & 1.22 \\
128 & 64 & 256 & 67.56 & 67.44 & 0.86 & 54.70 & 54.88 & 54.70 & 78.39 & 1.21 \\
\textbf{128} & \textbf{64} & \textbf{512} & \textbf{71.53} & \textbf{71.38} & \textbf{0.83} & \textbf{58.75} & \textbf{58.81} & \textbf{58.75} & \textbf{81.73} & \textbf{1.21} \\
128 & 64 & 1024 & 70.75 & 69.63 & 0.78 & 57.59 & 57.78 & 57.59 & 80.61 & 1.07 \\
128 & 96 & 256 & 70.84 & 70.69 & 0.83 & 57.27 & 57.27 & 57.27 & 79.93 & 1.25 \\
128 & 96 & 512 & 72.66 & 70.81 & 0.77 & 57.41 & 57.53 & 57.41 & 80.44 & 1.12 \\
128 & 96 & 1024 & 71.53 & 71.19 & 0.76 & 57.39 & 57.49 & 57.39 & 81.00 & 1.10 \\
128 & 120 & 256 & 71.22 & 69.88 & 0.80 & 56.21 & 56.25 & 56.21 & 78.98 & 1.15 \\
128 & 120 & 512 & 69.50 & 68.53 & 0.82 & 56.29 & 56.21 & 56.29 & 78.56 & 1.12 \\
128 & 120 & 1024 & 71.38 & 70.03 & 0.80 & 57.34 & 57.52 & 57.34 & 80.22 & 1.17 \\
256 & 128 & 256 & 68.94 & 68.25 & 0.86 & 55.38 & 55.44 & 55.38 & 78.80 & 1.28 \\
256 & 128 & 512 & 71.56 & 71.56 & 0.78 & 58.30 & 58.44 & 58.30 & 81.17 & 1.15 \\
\textbf{256} & \textbf{128} & \textbf{1024} & \textbf{72.38} & \textbf{72.16} & \textbf{0.74} & \textbf{57.95} & \textbf{58.00} & \textbf{57.95} & \textbf{80.98} & \textbf{1.11} \\
256 & 192 & 256 & 70.94 & 69.59 & 0.93 & 56.05 & 56.31 & 56.05 & 79.08 & 1.37 \\
256 & 192 & 512 & 72.19 & 70.56 & 0.78 & 58.14 & 58.25 & 58.14 & 80.60 & 1.12 \\
\textbf{256} & \textbf{192} & \textbf{1024} & \textbf{72.53} & \textbf{72.59} & \textbf{0.74} & \textbf{58.98} & \textbf{59.03} & \textbf{58.98} & \textbf{81.57} & \textbf{1.06} \\
256 & 240 & 256 & 69.81 & 67.94 & 0.82 & 54.84 & 54.92 & 54.84 & 77.82 & 1.16 \\
256 & 240 & 512 & 70.16 & 70.41 & 0.80 & 56.34 & 56.39 & 56.34 & 79.41 & 1.19 \\
256 & 240 & 1024 & 71.25 & 70.69 & 0.76 & 58.32 & 58.40 & 58.32 & 80.83 & 1.12 \\
\bottomrule
\end{tabular}
\end{table}

\begin{table}[H]
\centering
\caption{STFT Ablation BI2014b P300}
\label{tab:bi2014b_p300_results}
\footnotesize
\begin{tabular}{ccc|c|cc|cccccc}
\toprule
 & & & val & seen & seen & unseen & unseen & unseen & unseen & unseen & unseen \\
nperseg & noverlap & nfft & acc & acc & loss & acc & f1 & recall & auc & pr\_auc & loss \\
\midrule
32 & 16 & 32 & 82.14 & 15.40 & 1.13 & 16.67 & 28.50 & 99.67 & 54.10 & 18.75 & 1.18 \\
32 & 16 & 256 & 82.14 & 19.10 & 1.12 & 19.35 & 28.35 & 95.72 & 52.85 & 18.02 & 1.17 \\
\textbf{32} & \textbf{16} & \textbf{512} & \textbf{82.14} & \textbf{19.17} & \textbf{1.13} & \textbf{20.29} & \textbf{28.44} & \textbf{95.07} & \textbf{51.54} & \textbf{17.54} & \textbf{1.18} \\
32 & 24 & 32 & 82.00 & 24.08 & 1.11 & 22.92 & 28.05 & 90.13 & 53.30 & 19.84 & 1.16 \\
32 & 24 & 256 & 82.14 & 32.48 & 1.12 & 26.64 & 28.37 & 87.17 & 53.06 & 18.34 & 1.16 \\
32 & 24 & 512 & 82.21 & 44.32 & 1.12 & 38.71 & 27.50 & 69.74 & 51.80 & 18.82 & 1.16 \\
\textbf{32} & \textbf{30} & \textbf{32} & \textbf{82.14} & \textbf{16.27} & \textbf{1.12} & \textbf{17.38} & \textbf{28.48} & \textbf{98.68} & \textbf{54.68} & \textbf{20.63} & \textbf{1.17} \\
32 & 30 & 256 & 82.21 & 34.77 & 1.11 & 38.93 & 27.09 & 68.09 & 52.77 & 18.40 & 1.16 \\
64 & 32 & 64 & 82.14 & 15.47 & 1.13 & 16.89 & 28.49 & 99.34 & 52.02 & 17.31 & 1.18 \\
64 & 32 & 256 & 82.14 & 17.28 & 1.12 & 20.67 & 28.19 & 93.42 & 50.15 & 17.21 & 1.17 \\
64 & 32 & 512 & 82.07 & 27.98 & 1.12 & 22.26 & 28.09 & 91.12 & 52.85 & 17.96 & 1.17 \\
64 & 48 & 64 & 82.21 & 16.48 & 1.12 & 16.89 & 28.49 & 99.34 & 53.81 & 18.14 & 1.16 \\
64 & 48 & 256 & 82.14 & 28.18 & 1.12 & 32.57 & 28.65 & 81.25 & 53.09 & 19.12 & 1.16 \\
64 & 48 & 512 & 82.14 & 22.06 & 1.12 & 19.79 & 28.39 & 95.39 & 50.37 & 16.45 & 1.16 \\
64 & 60 & 64 & 82.14 & 15.74 & 1.11 & 18.91 & 28.38 & 96.38 & 54.22 & 19.93 & 1.16 \\
64 & 60 & 256 & 82.21 & 25.62 & 1.12 & 22.42 & 28.06 & 90.79 & 53.34 & 18.50 & 1.15 \\
64 & 60 & 512 & 82.14 & 19.30 & 1.12 & 16.67 & 28.57 & 100.00 & 50.34 & 16.71 & 1.15 \\
128 & 64 & 128 & 82.14 & 34.97 & 1.12 & 43.86 & 27.48 & 63.82 & 52.49 & 17.80 & 1.17 \\
128 & 64 & 256 & 81.87 & 24.21 & 1.12 & 26.54 & 29.32 & 91.45 & 54.09 & 18.71 & 1.16 \\
128 & 64 & 512 & 82.00 & 15.87 & 1.12 & 16.83 & 28.61 & 100.00 & 52.11 & 17.23 & 1.16 \\
128 & 96 & 128 & 80.30 & 15.60 & 1.12 & 17.00 & 28.52 & 99.34 & 52.84 & 17.59 & 1.15 \\
128 & 96 & 256 & 82.14 & 18.77 & 1.12 & 20.56 & 28.37 & 94.41 & 51.90 & 17.64 & 1.17 \\
128 & 96 & 512 & 82.14 & 23.87 & 1.13 & 37.12 & 27.54 & 71.71 & 51.31 & 17.38 & 1.18 \\
\textbf{128} & \textbf{120} & \textbf{128} & \textbf{82.14} & \textbf{15.40} & \textbf{1.13} & \textbf{16.67} & \textbf{28.57} & \textbf{100.00} & \textbf{54.67} & \textbf{19.83} & \textbf{1.18} \\
128 & 120 & 256 & 82.14 & 29.12 & 1.13 & 24.73 & 28.30 & 89.14 & 50.85 & 17.30 & 1.18 \\
128 & 120 & 512 & 82.14 & 20.91 & 1.12 & 18.20 & 28.61 & 98.36 & 48.39 & 16.50 & 1.15 \\
\bottomrule
\end{tabular}
\end{table}

\begin{table}[H]
\centering
\caption{STFT Ablation BNCI2014 P300}
\label{tab:bnci2014_p300_results}
\footnotesize
\begin{tabular}{ccc|c|cc|cccccc}
\toprule
 & & & val & seen & seen & unseen & unseen & unseen & unseen & unseen & unseen \\
nperseg & noverlap & nfft & acc & acc & loss & acc & f1 & recall & auc & pr\_auc & loss \\
\midrule
128 & 64 & 256 & 74.44 & 83.32 & 0.23 & 83.33 & 90.91 & 100.00 & 60.09 & 87.14 & 0.23 \\
128 & 64 & 512 & 72.83 & 83.61 & 0.23 & 83.64 & 91.05 & 99.86 & 59.84 & 86.38 & 0.23 \\
128 & 64 & 1024 & 72.87 & 83.32 & 0.23 & 83.33 & 90.91 & 100.00 & 58.68 & 86.10 & 0.23 \\
128 & 96 & 256 & 78.32 & 83.57 & 0.23 & 84.34 & 91.40 & 99.86 & 62.86 & 87.18 & 0.23 \\
128 & 96 & 512 & 78.16 & 83.40 & 0.23 & 83.60 & 91.03 & 99.88 & 64.72 & 88.37 & 0.23\\
\textbf{128} & \textbf{96} & \textbf{1024} & \textbf{71.92} & \textbf{82.99} & \textbf{0.23} & \textbf{84.41} & \textbf{91.42} & \textbf{99.61} & \textbf{65.19} & \textbf{88.60} & \textbf{0.22} \\
\textbf{128} & \textbf{120} & \textbf{256} & \textbf{74.03} & \textbf{83.69} & \textbf{0.22} & \textbf{83.95} & \textbf{91.21} & \textbf{99.88} & \textbf{65.05} & \textbf{89.14} & \textbf{0.22} \\
128 & 120 & 512 & 72.54 & 83.53 & 0.22 & 83.78 & 91.13 & 100.00 & 62.40 & 87.68 & 0.22 \\
128 & 120 & 1024 & 83.44 & 83.61 & 0.24 & 84.05 & 91.26 & 99.93 & 62.56 & 87.81 & 0.24 \\
\textbf{256} & \textbf{192} & \textbf{256} & \textbf{71.80} & \textbf{83.48} & \textbf{0.23} & \textbf{84.14} & \textbf{91.29} & \textbf{99.70} & \textbf{64.28} & \textbf{88.64} & \textbf{0.22} \\
256 & 192 & 512 & 77.95 & 83.36 & 0.23 & 83.31 & 90.90 & 99.95 & 60.43 & 86.58 & 0.23 \\
256 & 192 & 1024 & 75.19 & 83.32 & 0.23 & 83.33 & 90.91 & 100.00 & 55.78 & 85.53 & 0.23 \\
256 & 240 & 256 & 82.41 & 83.28 & 0.23 & 83.33 & 90.91 & 100.00 & 63.01 & 88.43 & 0.23 \\
256 & 240 & 512 & 81.92 & 83.28 & 0.23 & 83.35 & 90.91 & 99.95 & 63.81 & 88.15 & 0.23 \\
256 & 240 & 1024 & 82.91 & 83.32 & 0.23 & 83.33 & 90.91 & 100.00 & 61.28 & 87.20 & 0.23 \\
\bottomrule
\end{tabular}
\end{table}

\begin{table}[H]
\centering
\caption{STFT Ablation MI}
\label{tab:mi_results}
\footnotesize
\begin{tabular}{ccc|c|cc|ccccc}
\toprule
 & & & val & seen & seen & unseen & unseen & unseen & unseen & unseen \\
nperseg & noverlap & nfft & acc & acc & loss & acc & f1 & recall & auc & loss \\
\midrule
256 & 128 & 512 & 35.42 & 29.46 & 1.34 & 24.13 & 19.51 & 24.13 & 50.03 & 1.39 \\
256 & 128 & 1024 & 39.58 & 35.71 & 1.32 & 27.08 & 26.13 & 27.08 & 51.73 & 1.49 \\
256 & 128 & 2048 & 35.12 & 35.71 & 1.34 & 23.44 & 20.72 & 23.44 & 51.72 & 1.41 \\
256 & 192 & 512 & 35.42 & 33.63 & 1.37 & 26.91 & 21.55 & 26.91 & 51.57 & 1.40 \\
\textbf{256} & \textbf{192} & \textbf{1024} & \textbf{38.39} & \textbf{35.71} & \textbf{1.34} & \textbf{28.47} & \textbf{23.27} & \textbf{28.47} & \textbf{53.26} & \textbf{1.41} \\
256 & 192 & 2048 & 31.85 & 31.25 & 1.38 & 25.35 & 18.44 & 25.35 & 51.86 & 1.39 \\
256 & 240 & 512 & 37.50 & 31.55 & 1.36 & 25.17 & 23.32 & 25.17 & 51.86 & 1.42 \\
\textbf{512} & \textbf{256} & \textbf{512} & \textbf{40.77} & \textbf{37.50} & \textbf{1.31} & \textbf{27.43} & \textbf{26.84} & \textbf{27.43} & \textbf{53.99} & \textbf{1.45} \\
512 & 256 & 1024 & 35.71 & 32.74 & 1.37 & 28.30 & 19.11 & 28.30 & 51.12 & 1.42 \\
512 & 256 & 2048 & 34.23 & 30.95 & 1.37 & 24.83 & 16.85 & 24.83 & 50.66 & 1.43 \\
512 & 384 & 512 & 39.29 & 33.04 & 1.32 & 27.08 & 22.40 & 27.08 & 53.65 & 1.42 \\
\textbf{512} & \textbf{384} & \textbf{1024} & \textbf{33.93} & \textbf{33.04} & \textbf{1.37} & \textbf{28.82} & \textbf{20.34} & \textbf{28.82} & \textbf{52.73} & \textbf{1.39} \\
512 & 384 & 2048 & 34.82 & 30.36 & 1.37 & 25.17 & 17.59 & 25.17 & 52.45 & 1.40 \\
512 & 480 & 512 & 35.71 & 31.85 & 1.37 & 27.08 & 23.93 & 27.08 & 53.41 & 1.39 \\
512 & 480 & 1024 & 36.31 & 32.74 & 1.35 & 27.26 & 23.36 & 27.26 & 53.15 & 1.39 \\
1000 & 500 & 1024 & 35.71 & 29.76 & 1.34 & 26.39 & 22.38 & 26.39 & 51.52 & 1.45 \\
1000 & 500 & 2048 & 33.63 & 27.08 & 1.39 & 23.44 & 18.06 & 23.44 & 51.11 & 1.41 \\
1000 & 500 & 4096 & 34.23 & 28.57 & 1.37 & 25.69 & 20.95 & 25.69 & 52.13 & 1.39 \\
1000 & 750 & 1024 & 36.90 & 33.63 & 1.35 & 23.44 & 21.25 & 23.44 & 51.61 & 1.41 \\
1000 & 750 & 2048 & 36.90 & 32.44 & 1.35 & 26.04 & 24.53 & 26.04 & 49.74 & 1.49 \\
1000 & 750 & 4096 & 37.20 & 32.14 & 1.32 & 23.78 & 22.23 & 23.78 & 49.91 & 1.50 \\
1000 & 937 & 1024 & 34.82 & 33.93 & 1.37 & 24.83 & 23.70 & 24.83 & 50.73 & 1.39 \\
1000 & 937 & 2048 & 33.63 & 29.17 & 1.35 & 23.78 & 21.87 & 23.78 & 49.78 & 1.41 \\
\bottomrule
\end{tabular}
\end{table}

\begin{table}[H]
\centering
\caption{STFT Ablation Lee2019 MI}
\label{tab:lee2019_mi_results}
\footnotesize
\begin{tabular}{ccc|c|cc|ccccc}
\toprule
 & & & val & seen & seen & unseen & unseen & unseen & unseen & unseen \\
nperseg & noverlap & nfft & acc & acc & loss & acc & f1 & recall & auc & loss \\
\midrule
32 & 16 & 32 & 57.34 & 51.06 & 0.67 & 51.88 & 66.81 & 96.89 & 56.01 & 0.69 \\
32 & 16 & 256 & 55.75 & 51.31 & 0.68 & 51.46 & 66.23 & 95.18 & 55.16 & 0.69 \\
32 & 16 & 512 & 55.00 & 49.31 & 0.68 & 50.84 & 66.78 & 98.82 & 54.10 & 0.69 \\
\textbf{32} & \textbf{24} & \textbf{32} & \textbf{57.66} & \textbf{53.81} & \textbf{0.67} & \textbf{52.52} & \textbf{66.80} & \textbf{95.54} & \textbf{56.37} & \textbf{0.69} \\
32 & 24 & 256 & 55.47 & 50.41 & 0.68 & 51.20 & 66.42 & 96.54 & 54.53 & 0.69 \\
\textbf{32} & \textbf{30} & \textbf{32} & \textbf{56.28} & \textbf{53.50} & \textbf{0.67} & \textbf{52.32} & \textbf{66.30} & \textbf{93.79} & \textbf{56.47} & \textbf{0.69} \\
64 & 32 & 64 & 55.16 & 49.25 & 0.68 & 50.70 & 66.76 & 99.04 & 55.21 & 0.70 \\
64 & 32 & 256 & 55.44 & 49.28 & 0.69 & 51.14 & 66.78 & 98.21 & 54.55 & 0.70 \\
64 & 32 & 512 & 55.63 & 49.19 & 0.68 & 50.77 & 66.81 & 99.11 & 54.65 & 0.69 \\
64 & 48 & 64 & 56.53 & 51.13 & 0.68 & 51.54 & 66.48 & 96.11 & 55.95 & 0.69 \\
64 & 48 & 256 & 55.47 & 52.91 & 0.68 & 52.13 & 66.40 & 94.61 & 53.91 & 0.70 \\
64 & 48 & 512 & 55.44 & 49.06 & 0.68 & 50.64 & 66.75 & 99.07 & 53.78 & 0.69 \\
\textbf{64} & \textbf{60} & \textbf{64} & \textbf{55.75} & \textbf{53.09} & \textbf{0.68} & \textbf{52.48} & \textbf{66.67} & \textbf{95.04} & \textbf{54.93} & \textbf{0.69} \\
128 & 64 & 128 & 54.78 & 48.00 & 0.69 & 49.95 & 66.58 & 99.71 & 53.55 & 0.70 \\
128 & 64 & 256 & 55.69 & 48.56 & 0.69 & 50.34 & 66.63 & 99.18 & 53.87 & 0.70 \\
128 & 64 & 512 & 54.63 & 48.72 & 0.68 & 50.66 & 66.67 & 98.71 & 53.40 & 0.69 \\
128 & 96 & 128 & 56.47 & 50.31 & 0.67 & 51.04 & 66.63 & 97.75 & 55.14 & 0.69 \\
128 & 96 & 256 & 56.13 & 49.53 & 0.68 & 51.38 & 67.06 & 99.00 & 55.48 & 0.69 \\
128 & 96 & 512 & 55.78 & 49.06 & 0.68 & 50.55 & 66.67 & 98.93 & 55.06 & 0.69 \\
128 & 120 & 128 & 55.66 & 52.72 & 0.68 & 51.63 & 66.12 & 94.39 & 54.25 & 0.69 \\
128 & 120 & 256 & 55.28 & 49.41 & 0.69 & 51.16 & 66.83 & 98.39 & 53.22 & 0.69 \\
\bottomrule
\end{tabular}
\end{table}

\subsection{Fusion Ablation}
\subsubsection{Fusion Methods}
\label{apdx: fusion}

Here is a breakdown of each fusion method that we tried in the fusion ablation study. 

\textbf{Static Fusion}: The simplest baseline approach using fixed equal weighting of both streams. Provides a baseline assuming both modalities contribute equally, with no learnable fusion parameters.

\begin{equation}
    \mathbf{f} = \frac{1}{2}\mathbf{x}_s + \frac{1}{2}\mathbf{x}_t
\end{equation}

\begin{equation}
    \mathbf{f}_{\text{out}} = \text{BatchNorm}(\mathbf{f})
\end{equation}

\textbf{Global Attention Fusion}: Trial-level dynamic weighting that learns to adaptively balance the two streams based on the input. Allows the model to dynamically weight each stream per trial, enabling task-adaptive fusion. The residual connection improves gradient flow.

\begin{equation}
    \mathbf{c} = [\mathbf{x}_s \| \mathbf{x}_t] \in \mathbb{R}^{B \times 2D}
\end{equation}

\begin{equation}
    \mathbf{a} = \text{softmax}\left(\frac{\mathbf{W}_2 \cdot \text{ReLU}(\mathbf{W}_1 \mathbf{c} + \mathbf{b}_1) + \mathbf{b}_2}{\tau}\right) \in \mathbb{R}^{B \times 2}
\end{equation}

\begin{equation}
    \mathbf{f} = a_0 \cdot \mathbf{x}_s + a_1 \cdot \mathbf{x}_t
\end{equation}

\begin{equation}
    \mathbf{f}_{\text{out}} = \text{BatchNorm}\left(\mathbf{f} + \sigma(\alpha) \cdot \frac{\mathbf{x}_s + \mathbf{x}_t}{2}\right)
\end{equation}

where $\tau$ is a temperature parameter, $\alpha$ is a learnable residual scaling factor, $\mathbf{W}_1 \in \mathbb{R}^{D \times 2D}$, $\mathbf{W}_2 \in \mathbb{R}^{2 \times D}$, and $\|$ denotes concatenation.

\textbf{Spatial Attention Fusion}: Channel-level attention that computes different fusion weights for each EEG channel before global pooling. Different brain regions may benefit from different spectral/temporal balances. This allows the model to learn region-specific fusion strategies.

Let $\mathbf{X}_s \in \mathbb{R}^{B \times C \times D}$ be the per-channel spectral features (before averaging), and $\mathbf{x}_t \in \mathbb{R}^{B \times D}$ be the temporal features broadcast to $\mathbf{X}_t \in \mathbb{R}^{B \times C \times D}$.

\begin{equation}
    \mathbf{C}_{i} = [\mathbf{X}_{s,i} \| \mathbf{X}_{t,i}] \in \mathbb{R}^{B \times 2D}, \quad \forall i \in \{1, \ldots, C\}
\end{equation}

\begin{equation}
    \mathbf{a}_i = \text{softmax}\left(\frac{\text{MLP}(\mathbf{C}_i)}{\tau}\right) \in \mathbb{R}^{B \times 2}
\end{equation}

\begin{equation}
    \mathbf{F}_i = a_{i,0} \cdot \mathbf{X}_{s,i} + a_{i,1} \cdot \mathbf{X}_{t,i}
\end{equation}

\begin{equation}
    \mathbf{f}_{\text{out}} = \text{BatchNorm}\left(\frac{1}{C}\sum_{i=1}^{C} \mathbf{F}_i\right)
\end{equation}

\textbf{Gated Linear Unit (GLU) Fusion}: A gating mechanism that learns to suppress noise and control information flow. The gating mechanism can learn to suppress noisy or irrelevant feature dimensions while preserving informative ones, acting as a learned noise filter.

\begin{equation}
    \mathbf{c} = \mathbf{x}_s + \mathbf{x}_t
\end{equation}

\begin{equation}
    \mathbf{g} = \sigma\left(\mathbf{W}_g [\mathbf{x}_s \| \mathbf{x}_t] + \mathbf{b}_g\right) \in \mathbb{R}^{B \times D}
\end{equation}

\begin{equation}
    \mathbf{f}_{\text{out}} = \text{BatchNorm}(\mathbf{c} \odot \mathbf{g})
\end{equation}

where $\sigma$ is the sigmoid function, $\odot$ denotes element-wise multiplication, and $\mathbf{W}_g \in \mathbb{R}^{D \times 2D}$.

\textbf{Multiplicative Fusion}: Element-wise product of projected features to capture second-order interactions. Multiplicative interactions can capture correlations between the two streams that additive fusion cannot, enabling the model to learn feature conjunctions.

\begin{equation}
    \mathbf{f} = (\mathbf{W}_s \mathbf{x}_s + \mathbf{b}_s) \odot (\mathbf{W}_t \mathbf{x}_t + \mathbf{b}_t)
\end{equation}

\begin{equation}
    \mathbf{f}_{\text{out}} = \text{BatchNorm}(\text{Dropout}(\mathbf{f}))
\end{equation}

where $\mathbf{W}_s, \mathbf{W}_t \in \mathbb{R}^{D \times D}$.

\textbf{Low-Rank Bilinear Fusion}: Full bilinear interaction approximated with low-rank factorization for efficiency. Captures rich pairwise interactions between all feature dimensions while maintaining computational efficiency through the low-rank bottleneck.

A full bilinear model would compute $\mathbf{x}_s^\top \mathbf{W} \mathbf{x}_t$ with $\mathbf{W} \in \mathbb{R}^{D \times D \times D}$, which is computationally prohibitive. Instead, we use a low-rank approximation:

\begin{equation}
    \mathbf{z}_s = \mathbf{U}_s \mathbf{x}_s \in \mathbb{R}^{B \times R}
\end{equation}

\begin{equation}
    \mathbf{z}_t = \mathbf{U}_t \mathbf{x}_t \in \mathbb{R}^{B \times R}
\end{equation}

\begin{equation}
    \mathbf{z} = \mathbf{z}_s \odot \mathbf{z}_t \in \mathbb{R}^{B \times R}
\end{equation}

\begin{equation}
    \mathbf{f}_{\text{out}} = \text{BatchNorm}(\text{Dropout}(\mathbf{W}_{\text{out}} \mathbf{z} + \mathbf{b}_{\text{out}}))
\end{equation}

where $\mathbf{U}_s, \mathbf{U}_t \in \mathbb{R}^{R \times D}$ are low-rank projection matrices, $R \ll D$ is the rank (typically 8-32), and $\mathbf{W}_{\text{out}} \in \mathbb{R}^{D \times R}$.

\textbf{Cross-Attention Fusion}: Multi-head cross-attention allowing one stream to attend to the other, followed by global weighting. Cross-attention allows the spectral stream to selectively incorporate information from the temporal stream based on learned relevance, providing a more flexible fusion mechanism than simple weighting.

Step 1: Multi-Head Cross-Attention

For $H$ attention heads with head dimension $d_h = D/H$:

\begin{equation}
    \mathbf{Q} = \mathbf{W}_Q \mathbf{x}_s, \quad \mathbf{K} = \mathbf{W}_K \mathbf{x}_t, \quad \mathbf{V} = \mathbf{W}_V \mathbf{x}_t
\end{equation}

\begin{equation}
    \text{Attention}(\mathbf{Q}, \mathbf{K}, \mathbf{V}) = \text{softmax}\left(\frac{\mathbf{Q} \mathbf{K}^\top}{\tau\sqrt{d_h}}\right)\mathbf{V}
\end{equation}

\begin{equation}
    \mathbf{x}_{\text{attn}} = \mathbf{W}_O \cdot \text{Concat}(\text{head}_1, \ldots, \text{head}_H)
\end{equation}

Step 2: Global Weighting

\begin{equation}
    \mathbf{w} = \text{softmax}\left(\frac{\text{MLP}([\mathbf{x}_s \| \mathbf{x}_{\text{attn}}])}{\tau}\right) \in \mathbb{R}^{B \times 2}
\end{equation}

\begin{equation}
    \mathbf{f} = w_0 \cdot \mathbf{x}_s + w_1 \cdot \mathbf{x}_{\text{attn}}
\end{equation}

\begin{equation}
    \mathbf{f}_{\text{out}} = \text{BatchNorm}\left(\mathbf{f} + \sigma(\alpha) \cdot \frac{\mathbf{x}_s + \mathbf{x}_t}{2}\right)
\end{equation}

\subsubsection{Fusion Ablation Results}
\begin{table}[H]
\centering
\caption{Fusion Strategy Ablation on Wang 2016 SSVEP}
\label{tab:ssvep_fusion_ablation}
\resizebox{\columnwidth}{!}{%
\begin{tabular}{ll|ccccc}
\toprule
\textbf{STFT Config} & \textbf{Strategy} & \textbf{Val Acc} & \textbf{Unseen Acc} & \textbf{Unseen F1} & \textbf{Unseen Rec} & \textbf{Unseen AUC} \\
\midrule
\multirow{7}{*}{\shortstack[l]{nperseg=128\\noverlap=64\\nfft=256}}
& Static            & 71.510 & 68.109 & 68.419 & 68.109 & 96.504 \\
& Global Attn       & 73.932 & 69.311 & 69.490 & 69.311 & 97.114 \\
& Spatial Attn      & 69.943 & 64.663 & 64.718 & 64.663 & 95.877 \\
& GLU               & 76.211 & 68.990 & 69.355 & 68.990 & 96.537 \\
& Multiplicative    & 71.795 & 67.147 & 67.440 & 67.147 & 95.512 \\
& Bilinear          & 69.658 & 64.663 & 64.687 & 64.663 & 95.171 \\
& Cross Attn        & 73.362 & 68.349 & 68.863 & 68.349 & 96.155 \\
\midrule
\multirow{7}{*}{\textbf{\shortstack[l]{nperseg=256\\noverlap=128\\nfft=256}}}
& Static            & 72.222 & 68.990 & 69.393 & 68.990 & 96.580 \\
& \textbf{Global Attn} & \textbf{76.211} & \textbf{72.756} & \textbf{73.000} & \textbf{72.756} & \textbf{97.543} \\
& Spatial Attn      & 72.080 & 67.788 & 67.929 & 67.788 & 96.727 \\
& GLU               & 76.638 & 72.035 & 72.333 & 72.035 & 96.937 \\
& Multiplicative    & 71.083 & 69.471 & 69.721 & 69.471 & 95.530 \\
& Bilinear          & 70.513 & 68.750 & 69.210 & 68.750 & 95.647 \\
& Cross Attn        & 75.214 & 72.035 & 72.180 & 72.035 & 97.073 \\
\midrule
\multirow{7}{*}{\shortstack[l]{nperseg=256\\noverlap=128\\nfft=512}}
& Static            & 70.085 & 63.942 & 64.216 & 63.942 & 94.849 \\
& Global Attn       & 72.650 & 67.388 & 67.596 & 67.388 & 94.758 \\
& Spatial Attn      & 71.225 & 64.022 & 64.132 & 64.022 & 94.683 \\
& GLU               & 75.783 & 71.474 & 71.592 & 71.474 & 96.872 \\
& Multiplicative    & 72.365 & 66.026 & 66.262 & 66.026 & 95.219 \\
& Bilinear          & 67.379 & 62.981 & 63.480 & 62.981 & 93.359 \\
& Cross Attn        & 75.783 & 72.516 & 72.692 & 72.516 & 96.668 \\
\bottomrule
\end{tabular}%
}
\end{table}

\begin{table}[H]
\centering
\caption{Fusion Strategy Ablation on Lee2019 SSVEP}
\label{tab:lee2019_ssvep_fusion_ablation}
\resizebox{\columnwidth}{!}{%
\begin{tabular}{ll|ccccc}
\toprule
\textbf{STFT Config} & \textbf{Strategy} & \textbf{Val Acc} & \textbf{Unseen Acc} & \textbf{Unseen F1} & \textbf{Unseen Rec} & \textbf{Unseen AUC} \\
\midrule
\multirow{7}{*}{\shortstack[l]{nperseg=128\\noverlap=64\\nfft=512}}
& Static            & 94.906 & 85.964 & 85.955 & 85.964 & 97.716 \\
& Global Attn       & 94.906 & 83.250 & 83.245 & 83.250 & 96.912 \\
& Spatial Attn      & 94.906 & 84.821 & 84.800 & 84.821 & 97.363 \\
& GLU               & 95.219 & 85.214 & 85.244 & 85.214 & 97.534 \\
& Multiplicative    & 95.250 & 85.518 & 85.498 & 85.518 & 97.460 \\
& Bilinear          & 94.562 & 85.411 & 85.394 & 85.411 & 97.338 \\
& Cross Attn        & 94.938 & 85.250 & 85.243 & 85.250 & 97.227 \\
\midrule
\multirow{7}{*}{\textbf{\shortstack[l]{nperseg=256\\noverlap=128\\nfft=1024}}}
& Static            & 95.125 & 86.464 & 86.504 & 86.464 & 97.844 \\
& Global Attn       & 95.156 & 85.750 & 85.711 & 85.750 & 97.558 \\
& Spatial Attn      & 94.758 & 86.000 & 86.024 & 86.000 & 97.597 \\
& GLU               & 94.781 & 84.786 & 84.768 & 84.786 & 97.197 \\
& Multiplicative    & 94.844 & 85.714 & 85.687 & 85.714 & 97.573 \\
& \textbf{Bilinear}          & \textbf{95.000} & \textbf{86.679} & \textbf{86.637} & \textbf{86.679} & \textbf{97.783} \\
& Cross Attn        & 95.031 & 85.286 & 85.254 & 85.286 & 97.376 \\
\midrule
\multirow{7}{*}{\shortstack[l]{nperseg=256\\noverlap=192\\nfft=1024}}
& Static            & 94.281 & 85.714 & 85.696 & 85.714 & 97.467 \\
& Global Attn       & 94.938 & 85.893 & 85.893 & 85.893 & 97.500 \\
& Spatial Attn      & 94.156 & 85.375 & 85.353 & 85.375 & 96.641 \\
& GLU               & 94.969 & 85.518 & 85.493 & 85.518 & 97.329 \\
& Multiplicative    & 94.438 & 84.679 & 84.672 & 84.679 & 97.363 \\
& Bilinear          & 94.719 & 84.429 & 84.432 & 84.429 & 96.868 \\
& Cross Attn        & 94.531 & 82.929 & 82.925 & 82.929 & 96.564 \\
\bottomrule
\end{tabular}%
}
\end{table}

\begin{table}[H]
\centering
\caption{Fusion Strategy Ablation on BI2014b P300}
\label{tab:bi2014b_fusion_ablation}
\resizebox{\columnwidth}{!}{%
\begin{tabular}{ll|ccccc}
\toprule
\textbf{STFT Config} & \textbf{Strategy} & \textbf{Val Acc} & \textbf{Unseen Acc} & \textbf{Unseen F1} & \textbf{Unseen Rec} & \textbf{Unseen AUC} \\
\midrule
\multirow{7}{*}{\textbf{\shortstack[l]{nperseg=32\\noverlap=16\\nfft=512}}}
& Static            & 69.121 & 68.202 & 32.084 & 45.066 & 60.730 \\
& Global Attn       & 65.372 & 67.599 & 33.220 & 48.355 & 62.038 \\
& Spatial Attn      & 65.372 & 69.737 & 33.654 & 46.053 & 63.271 \\
& GLU               & 62.645 & 63.322 & 29.801 & 46.711 & 55.929 \\
& Multiplicative    & 65.167 & 66.118 & 25.721 & 35.197 & 57.451 \\
& \textbf{Bilinear} & \textbf{74.506} & \textbf{73.520} & \textbf{30.101} & \textbf{34.211} & \textbf{59.848} \\
& Cross Attn        & 64.963 & 66.338 & 27.078 & 37.500 & 59.078 \\
\midrule
\multirow{7}{*}{\shortstack[l]{nperseg=32\\noverlap=30\\nfft=32}}
& Static            & 67.144 & 66.228 & 28.205 & 39.803 & 58.639 \\
& Global Attn       & 69.325 & 71.382 & 29.839 & 36.513 & 60.829 \\
& Spatial Attn      & 60.123 & 59.978 & 28.988 & 49.013 & 58.069 \\
& GLU               & 63.258 & 62.555 & 29.515 & 47.039 & 57.759 \\
& Multiplicative    & 62.509 & 62.171 & 28.571 & 45.395 & 58.654 \\
& Bilinear          & 61.963 & 61.897 & 28.424 & 45.395 & 57.899 \\
& Cross Attn        & 64.894 & 61.952 & 28.454 & 45.395 & 58.056 \\
\midrule
\multirow{7}{*}{\shortstack[l]{nperseg=128\\noverlap=120\\nfft=128}}
& Static            & 72.529 & 72.423 & 30.042 & 35.526 & 60.018 \\
& Global Attn       & 65.235 & 64.748 & 27.508 & 40.132 & 57.483 \\
& Spatial Attn      & 62.849 & 62.664 & 28.691 & 45.066 & 59.455 \\
& GLU               & 62.781 & 66.118 & 34.255 & 52.961 & 62.957 \\
& Multiplicative    & 61.486 & 61.732 & 30.478 & 50.329 & 61.736 \\
& Bilinear          & 62.236 & 63.103 & 27.712 & 42.434 & 58.111 \\
& Cross Attn        & 63.122 & 66.118 & 28.966 & 41.447 & 60.058 \\
\bottomrule
\end{tabular}%
}
\end{table}

\begin{table}[H]
\centering
\caption{Fusion Strategy Ablation on BNCI2014-009 P300}
\label{tab:bnci2014_p300_fusion_ablation}
\resizebox{\columnwidth}{!}{%
\begin{tabular}{ll|ccccc}
\toprule
\textbf{STFT Config} & \textbf{Strategy} & \textbf{Val Acc} & \textbf{Unseen Acc} & \textbf{Unseen F1} & \textbf{Unseen Rec} & \textbf{Unseen AUC} \\
\midrule
\multirow{7}{*}{\shortstack[l]{nperseg=128\\noverlap=96\\nfft=1024}}
& Static            & 85.301 & 87.751 & 92.520 & 90.903 & 89.936 \\
& Global Attn       & 86.210 & 88.580 & 93.089 & 92.292 & 90.390 \\
& Spatial Attn      & 85.756 & 88.619 & 93.070 & 91.713 & 90.766 \\
& GLU               & 83.485 & 87.191 & 92.042 & 88.889 & 91.374 \\
& Multiplicative    & 86.581 & 89.410 & 93.624 & 93.310 & 90.953 \\
& Bilinear          & 86.664 & 89.275 & 93.529 & 93.009 & 90.647 \\
& Cross Attn        & 85.260 & 87.519 & 92.353 & 90.440 & 90.394 \\
\midrule
\multirow{7}{*}{\textbf{\shortstack[l]{nperseg=128\\noverlap=120\\nfft=256}}}
& Static            & 84.682 & 88.465 & 92.940 & 91.111 & 91.036 \\
& Global Attn       & 87.903 & 88.214 & 92.955 & 93.310 & 88.599 \\
& Spatial Attn      & 85.343 & 88.484 & 92.932 & 90.856 & 90.969 \\
& GLU               & 84.765 & 87.596 & 92.333 & 89.630 & 91.103 \\
& \textbf{Multiplicative} & \textbf{87.903} & \textbf{89.815} & \textbf{93.859} & \textbf{93.403} & \textbf{90.400} \\
& Bilinear          & 88.233 & 89.699 & 93.814 & 93.727 & 91.694 \\
& Cross Attn        & 86.168 & 89.178 & 93.427 & 92.292 & 91.685 \\
\midrule
\multirow{7}{*}{\shortstack[l]{nperseg=256\\noverlap=192\\nfft=256}}
& Static            & 86.788 & 89.352 & 93.587 & 93.241 & 91.220 \\
& Global Attn       & 84.393 & 86.748 & 91.791 & 88.912 & 90.294 \\
& Spatial Attn      & 83.237 & 86.632 & 91.715 & 88.796 & 89.918 \\
& GLU               & 86.210 & 89.063 & 93.372 & 92.454 & 91.126 \\
& Multiplicative    & 85.838 & 88.677 & 93.113 & 91.852 & 91.135 \\
& Bilinear          & 84.228 & 87.905 & 92.508 & 89.606 & 92.356 \\
& Cross Attn        & 85.136 & 87.654 & 92.321 & 89.051 & 91.523 \\
\bottomrule
\end{tabular}%
}
\end{table}

\begin{table}[H]
\centering
\caption{Fusion Strategy Ablation on BNCI2014-001 MI}
\label{tab:mi_fusion_ablation}
\resizebox{\columnwidth}{!}{%
\begin{tabular}{ll|ccccc}
\toprule
\textbf{STFT Config} & \textbf{Strategy} & \textbf{Val Acc} & \textbf{Unseen Acc} & \textbf{Unseen F1} & \textbf{Unseen Rec} & \textbf{Unseen AUC} \\
\midrule
\multirow{7}{*}{\shortstack[l]{nperseg=512\\noverlap=384\\nfft=1024}}
& Static            & 47.917 & 28.472 & 26.085 & 28.472 & 54.502 \\
& Global Attn       & 43.750 & 27.083 & 26.188 & 27.083 & 53.981 \\
& Spatial Attn      & 36.607 & 24.479 & 22.860 & 24.479 & 50.203 \\
& GLU               & 36.012 & 24.479 & 24.411 & 24.479 & 50.760 \\
& Multiplicative    & 47.321 & 28.299 & 27.440 & 28.299 & 55.923 \\
& Bilinear          & 50.000 & 28.993 & 28.586 & 28.993 & 56.170 \\
& Cross Attn        & 55.060 & 28.125 & 26.986 & 28.125 & 57.382 \\
\midrule
\multirow{7}{*}{\textbf{\shortstack[l]{nperseg=512\\noverlap=256\\nfft=512}}}
& Static            & 54.762 & 28.125 & 27.264 & 28.125 & 55.060 \\
& Global Attn       & 50.298 & 28.472 & 26.788 & 28.472 & 55.892 \\
& Spatial Attn      & 34.226 & 25.521 & 23.843 & 25.521 & 52.102 \\
& GLU               & 44.048 & 26.563 & 24.716 & 26.563 & 54.395 \\
& \textbf{Multiplicative} & \textbf{54.464} & \textbf{30.729} & \textbf{30.029} & \textbf{30.729} & \textbf{56.532} \\
& Bilinear          & 30.655 & 26.042 & 15.248 & 26.042 & 48.723 \\
& Cross Attn        & 50.000 & 27.778 & 23.785 & 27.778 & 58.496 \\
\midrule
\multirow{7}{*}{\shortstack[l]{nperseg=256\\noverlap=192\\nfft=1024}}
& Static            & 44.643 & 25.000 & 24.727 & 25.000 & 54.229 \\
& Global Attn       & 36.905 & 26.563 & 22.866 & 26.563 & 52.112 \\
& Spatial Attn      & 46.131 & 27.083 & 25.778 & 27.083 & 55.271 \\
& GLU               & 36.905 & 24.826 & 21.697 & 24.826 & 50.911 \\
& Multiplicative    & 52.083 & 29.688 & 28.651 & 29.688 & 57.098 \\
& Bilinear          & 50.893 & 29.514 & 27.590 & 29.514 & 59.252 \\
& Cross Attn        & 42.262 & 27.431 & 25.301 & 27.431 & 52.867 \\
\bottomrule
\end{tabular}%
}
\end{table}

\begin{table}[H]
\centering
\caption{Fusion Strategy Ablation on Lee2019 MI}
\label{tab:lee2019_mi_fusion_ablation}
\resizebox{\columnwidth}{!}{%
\begin{tabular}{ll|ccccc}
\toprule
\textbf{STFT Config} & \textbf{Strategy} & \textbf{Val Acc} & \textbf{Unseen Acc} & \textbf{Unseen F1} & \textbf{Unseen Rec} & \textbf{Unseen AUC} \\
\midrule
\multirow{7}{*}{\shortstack[l]{nperseg=32\\noverlap=24\\nfft=32}}
& Static            & 76.969 & 74.714 & 76.266 & 81.250 & 83.905 \\
& Global Attn       & 77.031 & 74.589 & 75.520 & 78.393 & 83.287 \\
& Spatial Attn      & 78.000 & 74.607 & 75.287 & 77.357 & 83.576 \\
& GLU               & 77.844 & 74.125 & 74.566 & 75.857 & 83.285 \\
& Multiplicative    & 78.344 & 75.500 & 77.149 & 82.714 & 84.523 \\
& Bilinear          & 77.625 & 74.214 & 72.620 & 68.393 & 84.000 \\
& Cross Attn        & 77.406 & 75.125 & 75.271 & 75.714 & 83.925 \\
\midrule
\multirow{7}{*}{\textbf{\shortstack[l]{nperseg=32\\noverlap=30\\nfft=32}}}
& Static            & 76.563 & 74.714 & 76.129 & 80.643 & 83.005 \\
& Global Attn       & 74.813 & 74.679 & 75.288 & 77.143 & 83.193 \\
& Spatial Attn      & 77.375 & 75.250 & 76.437 & 80.286 & 84.481 \\
& GLU               & 76.719 & 75.357 & 76.067 & 78.321 & 84.637 \\
& \textbf{Multiplicative} & \textbf{78.406} & \textbf{75.696} & \textbf{76.723} & \textbf{80.107} & \textbf{83.754} \\
& Bilinear          & 77.281 & 75.232 & 76.910 & 82.500 & 84.301 \\
& Cross Attn        & 75.375 & 74.714 & 76.870 & 84.036 & 83.780 \\
\midrule
\multirow{7}{*}{\shortstack[l]{nperseg=64\\noverlap=60\\nfft=64}}
& Static            & 76.313 & 74.429 & 75.293 & 77.929 & 83.844 \\
& Global Attn       & 77.281 & 73.286 & 75.777 & 83.571 & 83.165 \\
& Spatial Attn      & 77.781 & 75.071 & 75.806 & 78.107 & 84.327 \\
& GLU               & 77.656 & 75.357 & 75.958 & 77.857 & 84.013 \\
& Multiplicative    & 78.563 & 74.875 & 76.285 & 80.821 & 83.780 \\
& Bilinear          & 77.438 & 75.482 & 75.746 & 76.571 & 84.003 \\
& Cross Attn        & 76.906 & 74.946 & 75.356 & 76.607 & 84.132 \\
\bottomrule
\end{tabular}%
}
\end{table}

\subsection{Training Details}
To ensure reproducibility, we adopt a standardized training protocol across all evaluated tasks. The model is trained for a maximum of 100 epochs using the Adam optimizer with a batch size of 64. We utilize an initial learning rate of $3 \times 10^{-4}$ with a weight decay of $5 \times 10^{-4}$. The learning rate is managed by a \texttt{ReduceLROnPlateau} scheduler, which reduces the rate by a factor of 0.5 following 5 epochs of stagnant validation loss. Regularization is enforced through a global dropout rate of 0.3, a CNN-specific dropout of 0.25, and gradient norm clipping at a threshold of 1.0. Training terminates early if validation accuracy does not improve for 15 consecutive epochs. All experiments are conducted with a fixed random seed of 44, 36, and 10 to ensure deterministic behavior.


\end{document}

%% file: math_commands.tex
\usepackage{amsmath,amsfonts,bm}

\def\eqref#1{equation~\ref{#1}}

\def\1{\bm{1}}

\DeclareMathAlphabet{\mathsfit}{\encodingdefault}{\sfdefault}{m}{sl}
\SetMathAlphabet{\mathsfit}{bold}{\encodingdefault}{\sfdefault}{bx}{n}

